\documentclass{article}

\usepackage{graphicx}

\usepackage{times}
\usepackage{amsmath}
\usepackage{amssymb}
\usepackage{amsxtra, latexsym, mathrsfs}
\usepackage{pifont, stmaryrd}

\if@mathematic
   \def\bd#1{\ensuremath{\mathchoice
                     {\mbox{\boldmath$\displaystyle\mathbf{#1}$}}
                     {\mbox{\boldmath$\textstyle\mathbf{#1}$}}
                     {\mbox{\boldmath$\scriptstyle\mathbf{#1}$}}
                     {\mbox{\boldmath$\scriptscriptstyle\mathbf{#1}$}}}}
\else
   \def\bd#1{\ensuremath{\mathchoice
                     {\mbox{\boldmath$\displaystyle#1$}}
                     {\mbox{\boldmath$\textstyle#1$}}
                     {\mbox{\boldmath$\scriptstyle#1$}}
                     {\mbox{\boldmath$\scriptscriptstyle#1$}}}}
\fi

\def\tr{\operatorname{tr}}
\def\div{\operatorname{div}}
\def\vec{\operatorname{vec}}

\def\adj{\operatorname{adj}}
\def\p{\partial}

\title{Riemannian Level-set methods\\ for Tensor-Valued Data}

\author{Mourad Z\'era\"{\i} \& Maher Moakher\\
Laboratory for Mathematical and Numerical Modeling in 
Engineering Science\\ National Engineering School at Tunis\\ 
ENIT-LAMSIN, B.P. 37, 1002 Tunis Belv\'{e}d\`{e}re, Tunisia\\ 
{\tt mourad.zerai@gmail.com, maher.moakher@enit.rnu.tn}}

\begin{document}

\maketitle

\begin{abstract}
We present a novel approach for the derivation of PDEs modeling 
curvature-driven flows for matrix-valued data. This approach is based on 
the Riemannian geometry of the manifold of Symmetric Positive Definite 
Matrices $\mathcal{P}(n)$. The differential geometric attributes of 
$\mathcal{P}(n)$ $-$such as the bi-invariant metric, the covariant derivative 
and the Christoffel symbols$-$  allow us  to extend scalar-valued mean 
curvature and snakes methods to the tensor data setting. Since the data 
live on $\mathcal{P}(n)$, these  methods have the natural property of 
preserving positive definiteness of the initial data. 
Experiments on three-dimensional real DT-MRI data show that the proposed 
methods are highly robust.
\end{abstract}

\section{Introduction} \label{sec:intro}

With the introduction of diffusion tensor magnetic resonance imaging (DT-MRI)
\cite{basser94}, 
there has been an ever increasing demand on rigorous, reliable and robust
methods for the processing of tensor-valued data such as the estimation, 
filtering, regularization and segmentation. Many well established PDE-based 
methods used for the processing of scalar-valued data have been extended 
in various ways to the processing of multi-valued data such as vector-valued 
data and smoothly constrained data \cite{BC98,kimmel,Sap95,Sap96,SR96,sochen98}.
Recently, some efforts have been directed toward the extension of these
methods to tensor fields \cite{arsigny,feddern,deriche,lenglet,Moakher05,weickert,zm06}. 
The generalization of the methods used for scalar- and vector-valued data
to tensor-valued data is being pursued with mainly three formalisms: 
the use of geometric invariants of tensors like eigenvalues, determinant, 
trace; the generalization of Di Zenzo's concept of a structure tensor 
for vector-valued images to tensor-valued data; and recently, 
differential-geometric methods.

The aim of the present paper is to generalize the total variation (TV) flow, 
mean curvature motion (MCM), modified mean curvature flow and self snakes
to tensor-valued data such as DT-MRI. The key ingredient for these 
generalizations is the use of the Riemannian geometry of the space of 
symmetric positive-definite (SPD) matrices.
The remainder of this paper is organized as follows. In Section~\ref{sec:diffg}
we give a compilation of results that gives the differential geometry 
of the Riemannian manifold of symmetric positive-definite matrices. 
In Section~\ref{sec:def} we fix notation and recall some facts about 
immersions between Riemannian manifolds and their mean curvature. 
We explain in Section~\ref{sec:MRI} how to describe a DT-MR image by 
differential-geometric concepts. Section~\ref{sec:gc} is the key of our 
paper in which we extend several mean curvature-based flows for the denoising 
and segmentation from the scalar and vector setting to the tensor one. 
In Section~\ref{sec:exp} we present some numerical results. 

\section{Differential Geometry of $\mathcal{P}(n)$} \label{sec:diffg}

Positive-definite matrices are omnipresent in many engineering and physical 
contexts. They play important roles in various disciplines such as control 
theory, continuum mechanics, numerical analysis, covariance analysis, 
signal processing, etc. Recently, they gained an increasing attention 
within the diffusion tensor magnetic resonance imaging (DT-MRI) community 
as they are used as an encoding for the principal diffusion directions 
ans strengths in biological tissues. 

We here recall some differential-geometric facts about the space of symmetric 
positive-definite matrices that have been recently published by the authors.
We denote by $\mathcal{S}(n)$ the vector space of $n\times n$ symmetric 
matrices. A matrix $A\in \mathcal{S}(n)$ is said to be positive semidefinite 
if $\bd{x}^{T}A \bd{x} \ge 0$ for all $\bd{x} \in \mathbb{R}^n$, and positive 
definite if in addition $A$ is invertible. The space of all $n \times n$ 
symmetric, positive-definite matrices will be denoted
by $\mathcal{P}(n)$. We note that the set of positive-semidefinite matrices 
is a pointed convex cone in the linear space of $n \times n$ matrices, and
that $\mathcal{P}(n)$ is the interior of this cone. It is a differentiable 
manifold endowed with a Riemannian structure.
The tangent space to $\mathcal{P}(n)$ at any of its points $P$ is the space 
$T_{P}\mathcal{P}(n)= \{P\}\times \mathcal{S}(n)$, which for simplicity is
identified with $\mathcal{S}(n)$. On each tangent space $T_P\mathcal{P}(n)$ 
we introduce the base point-dependent inner product defined by 
$\langle A, B \rangle_P := \tr(P^{-1} A P^{-1} B)$.

This inner product leads to a natural Riemannian metric on the manifold 
$\mathcal{P}(n)$ that is given at each $P$ by the differential
\begin{equation} \label{eq:metric}
ds^2 = \tr\left(P^{-1} dP P^{-1} dP \right),
\end{equation}
where $dP$ is the symmetric matrix with elements $(dP_{ij})$.
We note that the metric (\ref{eq:metric}) is invariant under 
congruent transformations: $P \rightarrow LPL^{T}$ and under inversion
$P \rightarrow P^{-1}$.

For an $n \times n$ matrix $A$ we denote by $\vec A$ the $n^2$-column vector
that is obtained by stacking the columns of $A$. If $A$ is symmetric, then
then $\tfrac12 n(n-1)$ elements of $\vec(A)$ are redundant. We will denote by 
$\upsilon(A)$ the $d=\tfrac12 n(n+1)$-vector that is obtained from $\vec(A)$ 
by eliminating the redundant elements, e.g., all supradiagonal elements of $A$.
We note that there are several ways to arrange the independent elements of 
$\vec(A)$ into $\upsilon(A)$. In any case, there exists a unique 
$n^{2} \times \tfrac12 n(n+1)$ matrix, called the duplication matrix
and denoted by $D_n$, that by duplicating certain elements, reconstructs 
$\vec A$ from $\upsilon(A)$, i.e., is the matrix such that
\begin{equation}
\vec A = D_n \upsilon(A).
\label{eq:defD}
\end{equation}
The duplication matrix $D_n$, which has been studied extensively by Henderson 
and Searle \cite{henderson79}, and by Magnus and Neudecker \cite{magnus80},
has full column rank $\tfrac{1}{2}n(n+1)$. Hence, 
$D_n^{T}D_n$ is non-singular and it follows that the duplication matrix $D_n$ 
has a Moore-Penrose inverse denoted by $D_n^{+}$ and is given by
\[
D_n^{+}=\left(D_n^{T}D_n \right)^{-1}D_n^T.
\]
It follows from (\ref{eq:defD}) that
\begin{equation}
\upsilon(A) = D_n^+ \vec A.
\label{eq:defD+}
\end{equation}

By using the vector $\upsilon(P)$ as a parametrization of 
$P \in \mathcal{P}(n)$ we obtain the matrix of components of the metric 
tensor associated with the Riemannian metric (\ref{eq:metric}) is given 
explicitly by \cite{zm06}
\begin{equation} \label{eq:G}
G(P) = D_n^{T}(P^{-1}\otimes P^{-1})D_n.
\end{equation}
For differential-geometric operators on $\mathcal{P}(n)$ it is important 
to obtain the expression of the inverse of the metric and that of its 
determinant. The matrix of components of the inverse metric tensor is given by
\begin{equation} \label{eq:G-1}
G^{-1}(P) = D_n^{+}\left(P \otimes P\right){D_n^{+}}^T ,
\end{equation}
and the determinant of $G$ is
\begin{equation} \label{eq:detG}
\det(G(P)) = 2^{n(n-1)/2}\left((\det(P)\right)^{(n+1)}.
\end{equation}

In the coordinate system $(p^\alpha)$, the Christoffel symbols are given by \cite{zm06}
\[
\Gamma^\gamma_{\alpha \beta} = - [D_n^T\left(P^{-1}\otimes E^\gamma\right)
D_n]_{\alpha \beta}, \quad 1\le \alpha, \beta, \gamma \le d,
\]
where $E^\gamma$ is the dual basis associated with the local coordinates
$(p^\alpha)$.
As the elements of $E^\gamma$ and $D_n$ are either 0, 1, or $\tfrac12$, 
it follows from the above theorem that each non-vanishing 
Christoffel symbol is given by an element of $P^{-1}$ or half of it.   

Let $P$ be an element of $\mathcal{P}(3)$ and let $dP$ be a (symmetric) 
infinitesimal variation of it
\[
P = \begin{bmatrix} p^1 & p^4 & p^6 \\\noalign{\smallskip} 
p^4 & p^2 & p^5 \\\noalign{\smallskip} p^6 & p^5 & p^3
\end{bmatrix}, 
\qquad
dP = \begin{bmatrix} dp^1 & dp^4 & dp^6 \\\noalign{\smallskip}
dp^4 & dp^2 & dp^5\\\noalign{\smallskip}
dp^6 & dp^5 & dp^3\end{bmatrix}. 
\]
Hence, the complete and reduced vector forms of $P$ are respectively,
\begin{equation*}
\vec (P) = [p^{1} \ p^4 \ p^6 \ p^4 \ p^{2} \ p^5 \
p^{6} \ p^{5} \ p^{3} ]^T, \quad 
\upsilon(P)  =  [p^{1} \ p^{2} \ p^{3} \ p^{4} \ p^{5} \ p^{6}]^T.
\end{equation*}
The components of the inverse metric tensor and the Christoffel symbols are 
given explicitly in the appendix.

\section{Immersions and Mean Curvature}\label{sec:def}

Let $(M,\gamma )$ and $(N, g)$ be two connected Riemanian manifolds of 
dimensions $m$ and $n$, respectively. 
We consider a map $\phi : M \rightarrow N$ that is of class $C^2$, i.e., 
$\phi \in C^2(M,N)$. Let $\{x^\alpha \}-{1\leq \alpha \leq m}$ be a local 
coordinate system of $x$ in a neighborhood of a point $p\in M$ and let 
$\{y^i\}_{1\leq i \leq n}$ be a local coordinate system of $y$ in a 
neighborhood of $\phi(P)\in N$. 

The mapping $\phi$ induces a metric $\phi^*g$ on $M$ defined by
\begin{equation}
\phi^*g\left(X_p,Y_p\right)=g\left(\phi_*(X_p),\phi_*(Y_p)\right).
\end{equation}
This metric is called the {\it pull-back} metric induced by $\phi$, 
as it maps the metric in the opposite direction of the mapping $\phi$.

An isometry is a diffeomorphism $\phi : M \rightarrow N$ that preserves the 
Riemannian metric, i.e., if $g$ and $\gamma$ are the metrics for $M$ and $N$, 
respectively, then $\gamma=\phi^*g$. It follows that an isometry preserves
the length of curves, i.e., if $c$ is a smooth curve on $M$, then the curve 
$\phi\circ c$ is a curve of the same length on $N$.
Also, the image of a geodesic under an isometry is again a geodesic.

A mapping $\phi : M \rightarrow N$ is called an {\it immersion} if 
$\left(\phi_*\right)_p$ is injective for every point $p$ in $M$.
We say that $M$ is immersed in $N$ by $\phi$ or that $M$ is an immersed 
submanifold of $N$. When an immersion $\phi$ is injective, it is called 
an {\it embedding} of $M$ into $N$. We then say that $M$ is an
{\it embedded submanifold}, or simply, a {\it submanifold} of $N$.

Now let $\phi : M \rightarrow N$ be an immersion of a manifold $M$ into a 
Riemannian manifold $N$ with metric $g$.
The first fundamental form associated with the immersion $\phi$ is 
$h = \phi^*g$. Its components are $h_{\alpha\beta} = \p_\alpha\phi^i
\p_\beta\phi^jg_{ij}$ where 
$\p_\alpha\phi^i=\frac{\p\phi^i}{\p x^\alpha}$. 
The total covariant derivative $\nabla d\phi$ is called the second 
fundamental form of $\phi$ and is denoted by $II^M(\phi)$.
The second fundamental form  $II^M$ takes values in the normal bundle of $M$.
The mean curvature vector $\bd{H}$ of an isometric immersion 
$\phi : M\rightarrow N$ is defined as the trace of the second fundamental 
form $II^M(\phi)$ divided by $m = \dim M$ \cite{jost}
\begin{equation} \label{mc}
\bd{H} := \frac1m \tr_\gamma II^M(\phi).
\end{equation}
In local coordinates, we have \cite{jost}
\begin{equation} \label{tf}
m{H}^i = \Delta_M \phi^i + \gamma^{\alpha\beta}(x)
^N\Gamma^i_{jk}\left(\phi(x)\right)\frac{\partial \phi^j}{\partial x^\alpha}
\frac{\partial \phi^k}{\partial x^\beta}.
\end{equation}
where $^N\Gamma^i_{jk}$ are the Christoffel symbols of $\left({N},g\right)$ 
and $\Delta_{M}$ is the Laplace-Beltrami operator on $(M,\gamma)$ given by
\begin{equation}
\Delta_M = \frac{1}{\sqrt{\det \gamma}}\frac{\partial}{\partial x^\alpha} 
\left( \sqrt{\det \gamma}\gamma^{\alpha\beta}
\frac{\partial }{\partial x^\beta}\right).
\end{equation}

\section{Diffusion-Tensor MRI Data as Isometric Immersions}\label{sec:MRI}

A volumetric tensor-valued image can be described mathematically as an 
isometric immersion $(x^1, x^2, x^3) \mapsto \phi = (x^1, x^2, x^3; 
P(x^1, x^2, x^3))$ of a three-dimensional domain $\Omega$ in the fiber 
bundle $\mathbb{R}^3\otimes \mathcal{P}(3)$, which is a nine-dimensional 
manifold. We denote by $(M,\gamma)$ the image manifold and its metric and 
by $(N,g)$ the target manifold and its metric. Here $M=\Omega$ and 
$N=\mathbb{r}^3\otimes \mathcal{P}(3)$. 
Consequently, a tensor-valued image is a section of this fiber bundle. 
The metric $\hat{g}$ of $N$ is given by
\begin{equation} \label{ds2}
d\hat s^2=ds^2_{\rm spatial} + ds^2_{\rm tensor}.
\end{equation}

The target manifold $N$, in this context is also called the
{\it space-feature manifold\/} \cite{sochen98}. We can rewrite the metric 
defined by (\ref{ds2}) as the quadratic form
\[
d\hat{s}^2 = (dx^1)^2 + (dx^2)^2 + (dx^3)^2 +
(d \bd{p})^T D_n^{T}(P^{-1}\otimes P^{-1})D_n (d \bd{p}),
\]
where $\bd{p} = (p^i) = \upsilon(P)$.
The corresponding metric tensor is
\[
\hat{g} = \begin{pmatrix} I_3 & 0_{3,6} \\ 0_{6,3} & g \end{pmatrix},
\]
where $g$ is the metric tensor of $\mathcal{P}(3)$ as defined in 
Section~\ref{sec:diffg}.

Since the image is an isometric immersion, we have $\gamma = \phi^*\hat{g}$.
Therefore
\begin{equation}
\gamma_{\alpha\beta} = \delta_{\alpha\beta} + g_{ij}\p_\alpha p^i\p_\beta
p^j, \quad \alpha,\beta =1,\ldots,m, \quad i,j=1,\ldots,d.
\end{equation}
We note that $d=n-m$ is the codimension of $M$. In compact form, we have
\begin{equation}
\gamma = I_m + (\nabla \bd{p})^T G(\phi) \nabla \bd{p}.
\end{equation}
where $G$ is given by (\ref{eq:G}). (We take $m=2$ for a slice 
and $m=3$ for a volumetric DT-MRI image.)

\section{Geometric Curvature-Driven Flows for Tensor-Valued Data}
\label{sec:gc}

The basic concept in which geometric curvature-driven flows are based is 
the mean curvature of a submanifold embedded in a higher dimensional manifold. 
Here we generalize the scalar mean curvature flow to mean curvature flow 
in the space-feature manifold $\Omega \otimes \mathcal{P}(3)$. For this, 
we embed the Euclidean image space $\Omega$ into the Riemannian manifold 
$\Omega \otimes \mathcal{P}(3)$, and use some classical results from 
differential geometry to derive the {\it Riemannan Mean Curvature} (RMC). 
We then use the RMC to generalize mean curvature flow to the tensor-valued 
data.
Given the expression of the mean curvature vector $\bd{H}$, we can 
establish some PDEs based tensor-image filtering. Especially, we are 
interested of the so called level-set methods, which relay on PDEs that 
modify the shape of level sets in an image.

\subsection{Riemannian Total Variation Flow}

The total variation norm (TV) method introduced in \cite{ROF92} and its
reconstructions have been successfully used in reducing noise and blurs 
without smearing sharp edges in grey-level, color and other vector-valued
images \cite{Cum91,LC91,Nev77,Sap95,Sap96,SR96}. It is then natural to 
look for the extension of the TV norm to tensor-valued images.

The TV norm method is obtained as a gradient-decent flow associated with
the $L^1$-norm of the tensor field. This yields the following PDE
that express the motion by the mean curvature vector $\bd{H}$
\begin{equation} \label{eq:TV-flow}
\p_t\phi^i = {H}^i.
\end{equation}
This flow can be considered as a deformation of the tensor field toward 
{\it minimal immersion}. Indeed, it derives from variational setting 
that minimize the {\it volume} of the embedded image manifold in the 
space-feature manifold.

\subsection{Riemannian Mean Curvature Flow}

The following flow was proposed for the processing of scalar-valued images
\begin{equation} \label{eq:s-mcf}
\p_tu=  |\nabla u|\div  \frac{\nabla u}{|\nabla u|},\quad 
u(0,x,y)=u_0(x,y),
\end{equation}
where $u_0(x,y)$ is the grey level of the image to be processed, 
$u(t,x,y)$ is its smoothed version that depends on the scale parameter $t$.

The ``philosophy" of this flow is that the term $|\nabla u| \div 
\frac{\nabla u}{|\nabla u|}$ represents a degenerate diffusion term 
which diffuses $u$ in the direction orthogonal to its gradient $\nabla u$ 
and does not diffuse at all in the direction of $\nabla u$.

This formulation has been proposed as a ``morphological scale space" 
\cite{Alvarez} and  as more numerically tractable method of solving total 
variation \cite{Marquina}.

The natural generalization of this flow to tensor-valued data is
\begin{equation} \label{eq:t-mcf}
\p_t\phi^i = |\nabla^\gamma\phi|_g H^i,\quad i=1,\ldots,d.
\end{equation}
where
\begin{equation*}
|\nabla^\gamma\phi|_g=\gamma^{\alpha\beta}g_{ij}\p_\alpha\phi^i\p_\beta\phi^j.
\end{equation*}
We note that several authors have tried to generalize curvature-driven
flows for tensor-valued data in different ways. We think that the 
the use of differential-geometric tools and concepts yield the correct
generalization.

\subsection{Modified Riemannian Mean Curvature Flow}

To denoise highly degraded images, Alvarez et al. \cite{Alvarez92}
have proposed a modification of the mean curvature flow equation 
(\ref{eq:s-mcf}) that reads
\begin{equation}
\p_t\phi= c\left(|K \star \nabla \phi |\right) |\nabla\phi|\div  \frac{\nabla\phi}{|\nabla\phi|},\quad \phi(0,x,y)=\phi_0(x,y),
\label{mmcf}
\end{equation}
where $K$ is a smoothing kernel (a Gaussian for example), $K \star \nabla\phi$ 
is therefore a local estimate of $\nabla\phi$ for noise elimination, and 
$c(s)$ is a nonincreasing real function which tends to zero as 
$s\rightarrow\infty$. We note that for the numerical experiments we have used $c(|\nabla\phi|)=k^2/(k^2+|\nabla\phi|^2)$.

The generalization of the modified mean curvature flow to tensor-field 
processing is
\begin{equation} \label{eq:MRMCF}
\p_t\phi^i= c\left(|K \star \nabla^\gamma \phi |_g\right) |\nabla^\gamma\phi|_g
H^i,\quad \phi^i(0,\Omega)=\phi^i_0(\Omega),
\end{equation}
The role of $c$ is to reduce the magnitude of smoothing near edges. 
In the scalar case, this equation does not have the same action as the 
Perona-Malik equation of enhancing edges. Indeed, Perona-Malik equation 
has variable diffusivity function and has been shown to selectively produce 
a ``negative diffusion" which can increase the contrast of edges. Equation 
of he form (\ref{mmcf}) have always positive or forward diffusion, and the 
term $c$ merely reduces the magnitude of that smoothing. To correct
this situation, Sapiro have proposed the self-snakes formalism 
\cite{Sap95}, which we present in the next subsection and generalize to the 
matrix-valued data setting.

\subsection{Riemannan Self-Snakes}

The method of Sapiro, which he names {\it self-snakes} introduces an 
edge-stopping function into mean curvature flow
\begin{equation}
\begin{array}{rcl}
  \p_t\phi & = &  |\nabla\phi|\div \left(c\left(K \star |\nabla\phi |\right)\frac{\nabla\phi}{|\nabla\phi|}\right) \\
           & = &  c\left(K \star |\nabla \phi |\right)|\nabla\phi|\div \left(\frac{\nabla\phi}{|\nabla\phi|}\right)+\nabla c\left(K \star |\nabla \phi|\right)\cdot\nabla\phi
\end{array}
\label{rss}
\end{equation}
Comparing equation (\ref{rss}) to (\ref{mmcf}), we observe that the term 
$\nabla c\left(K \star |\nabla \phi|\right)\cdot\nabla\phi$ is missing in 
the old model. This is due to the fact that the Sapiro model takes into 
account the image structure. Indeed, equation (\ref{rss}) can be re-written as
\begin{equation} \label{ds}
\p_t \phi = \mathcal{F}_{\rm diffusion}+\mathcal{F}_{\rm shock},
\end{equation}
where
\[
\mathcal{F}_{\rm diffusion}=c\left(K \star |\nabla \phi |\right)|\nabla\phi|
\div \left(\frac{\nabla\phi}{|\nabla\phi|}\right),
\]
\[
\mathcal{F}_{\rm shock}=\nabla c\left(K \star |\nabla \phi|\right)\cdot
\nabla\phi.
\]
The term $\mathcal{F}_{\rm diffusion}$ is as in the anisotropic flow 
proposed in \cite{Alvarez92}. The second term in (\ref{ds}), i.e., 
$\nabla c\cdot\nabla\phi$, increases the attraction of the deforming 
contour toward the boundary of ``objects'' acting as the shock-filter 
introduced in \cite{Osher} for deblurring. Therefore,
the flow $\nabla c\cdot\nabla\phi$ is a shock filter acting like the backward 
diffusion in the Perona-Malik equation, which is
responsible for the edge-enhancing properties of self snakes. 
See \cite{Sap95} for detailed discussion on this topic.

We are now interested in generalizing Self-Snakes method for the case of 
tensor-valued data. We will start the generalization
from equation (\ref{ds}) in the following manner
\begin{equation} \label{eq:RSSF}
\p_t \phi = \mathcal{F}_{\it diffusion}+\mathcal{F}_{\it shock},
\end{equation}
where
\begin{equation}
\begin{array}{lcl}
\mathcal{F}_{\it diffusion}  & = & c\left(K \star |\nabla^\gamma \phi |_g
\right)|\nabla^\gamma\phi|_g {H}^i \\
\mathcal{F}_{\it shock} & = & \nabla c\left(K \star |\nabla^\gamma \phi|_g
\right)\cdot\nabla^\gamma\phi^i.
\end{array}
\end{equation}
This decomposition is not artificial, since the covariant derivative on 
follow the same chain rule as the Euclidean directional derivative: let $V$ 
a vector field on $M$ which components are $v^i$, and let $\rho$ a scalar 
function. From the classic differential geometry we have
\begin{equation}
\nabla^\gamma_i(\rho v^i)=\rho\nabla^\gamma_iv^i+v^i\nabla^\gamma_i\rho
\end{equation}
and in compact form
\begin{equation}
\div _\gamma(\rho V)=\rho\div_\gamma V+V\cdot\mbox{grad}_\gamma\rho.
\end{equation}

\section{Numerical Experiments}  \label{sec:exp}

In Fig.~1 (left), we give a slice of a 3D tensor field defined over a square in
$\mathbb{R}^2$. We note that a symmetric positive-definite $3 \times 3$ 
matrix $P$ is represented graphically by an ellipsoid whose principal 
directions are parallel to the eigenvectors of $P$ and whose axes are 
proportional to the eigenvalues of $P^{-1}$. Figure~1 (right) shows this tensor 
field after the addition of noise. The resultant tensor field 
$P_0(x^1,x^2,x^3)$ is used as an initial condition for the partial 
differential equations (\ref{eq:RSSF})
which we solve by a finite difference scheme with Neumann
boundary conditions. We used 50 time steps of 0.01.s. Figure~2 represents the tensor
smoothed by (\ref{eq:RSSF}).

\begin{figure}[h!]
\begin{center}
\includegraphics[width=.45\textwidth]{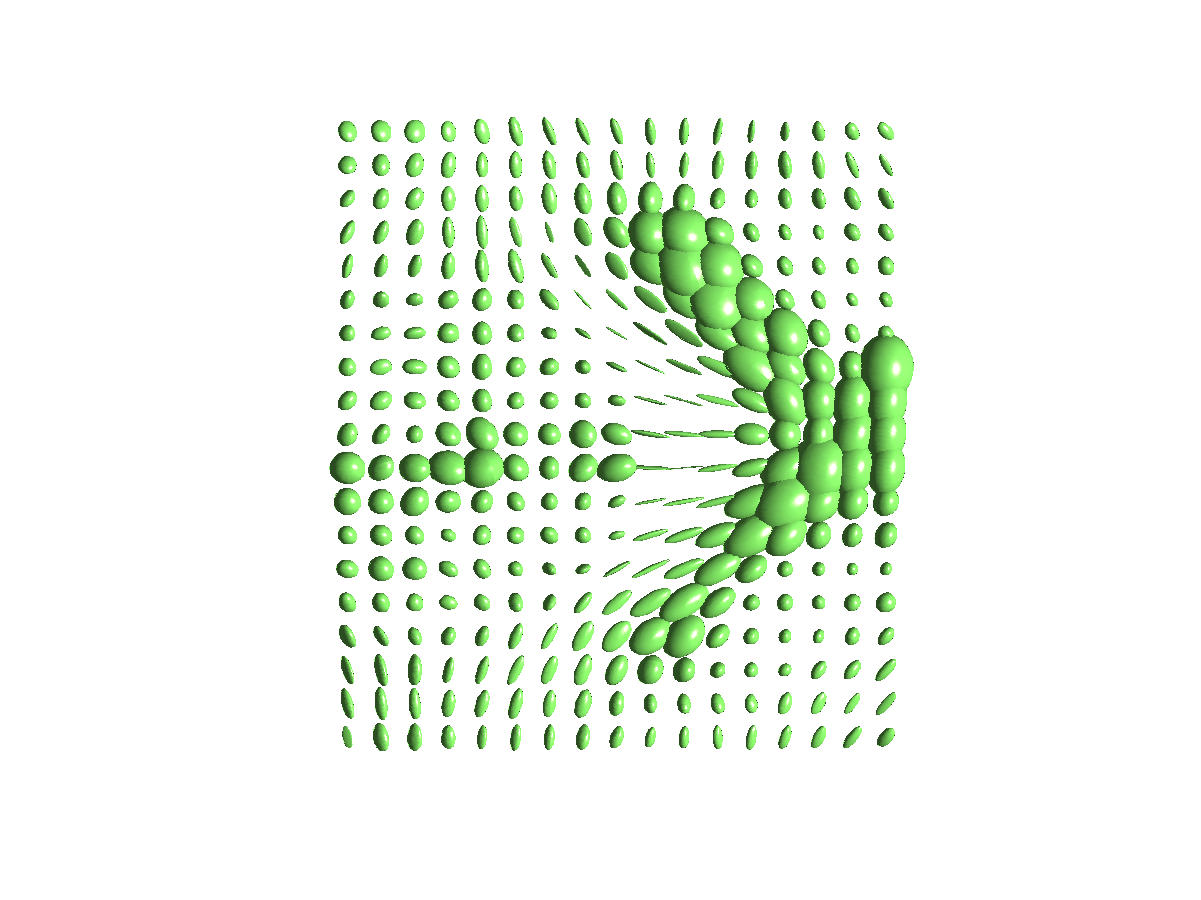} \quad
\includegraphics[width=.45\textwidth]{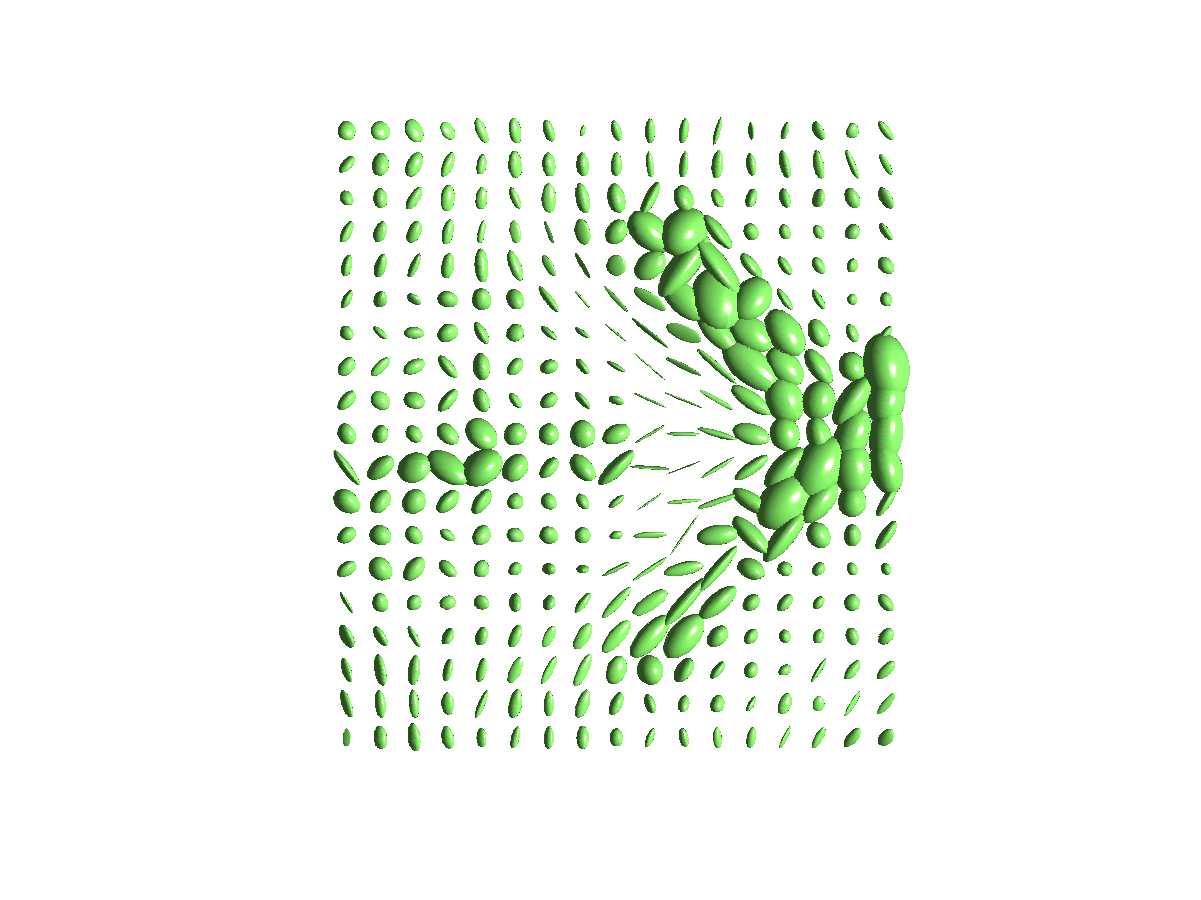}
\caption{Original tensor field (left)and noisy tensor field 
(right).}
\end{center}
\end{figure}

\begin{figure}[h!]
\begin{center}
\includegraphics[width=.45\textwidth]{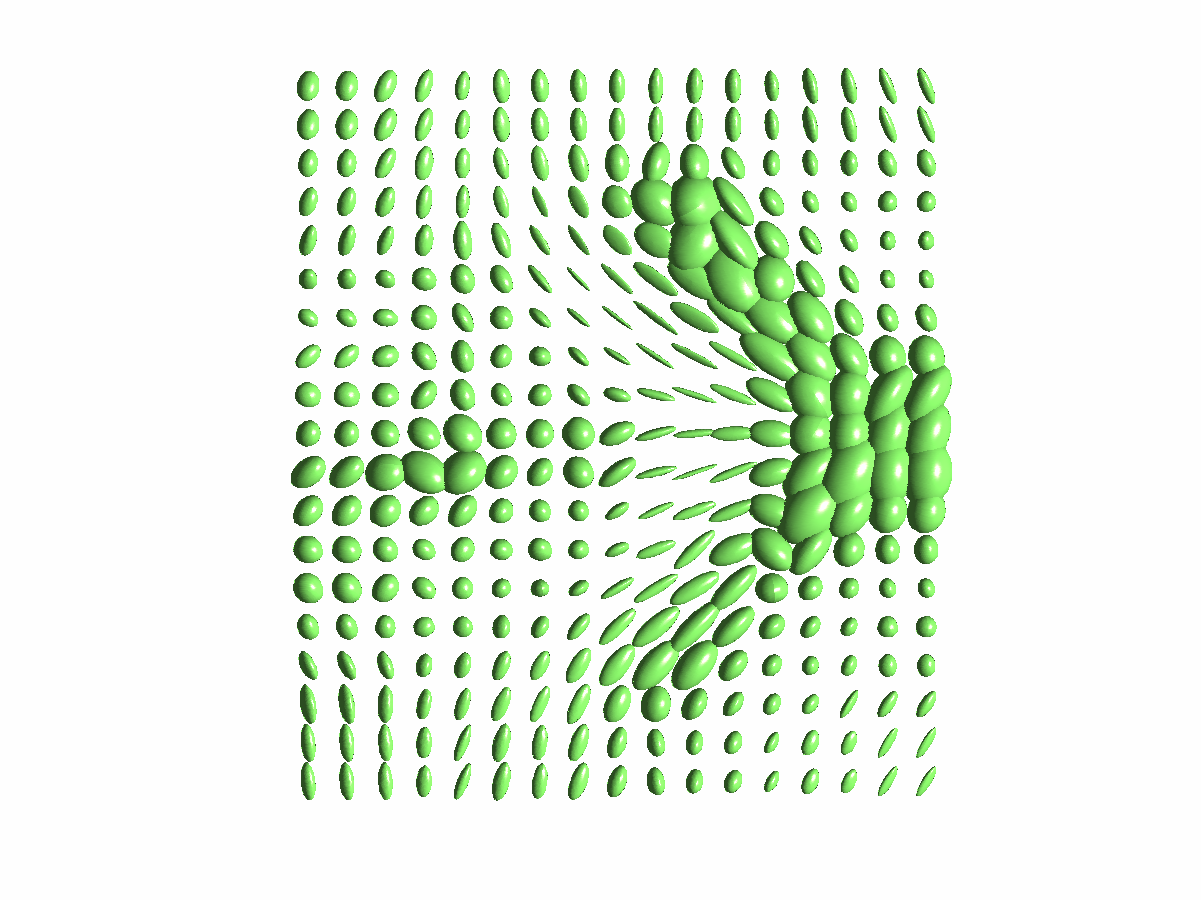}
\caption{Tensor field smoothed by the Riemannian self snake flow.}
\end{center}
\end{figure}

In this paper we generalized several curvature-driven flows of scalar-
and vector-valued data to tensor-valued data. The use
of the differential-geometric tools and concepts yields a natural 
extension of these well-known scalar-valued data processing methods 
to tensor-valued data processing.


\section*{Appendix}

We give here the explicit form of the inverse metric tensor and
Christoffel symbols for the Riemannian metric on $\mathcal{P}(3)$.
The components of the inverse metric tensor are given by
\begin{equation*}
G^{-1} = \begin{bmatrix} 
(p^1)^2 & (p^4)^2 & (p^6)^2 & p^1p^4 & p^4p^6 & p^1p^6\\\noalign{\smallskip}
(p^4)^2 & (p^2)^2 & (p^5)^2 & p^2p^4 & p^2p^5 & p^4p^5\\\noalign{\smallskip}
(p^6)^2 & (p^5)^2 & (p^3)^2 & p^6p^5 & p^5p^3 & p^6p^3\\\noalign{\smallskip}
p^1p^4 & p^2p^4 & p^6p^5 &  \tfrac12 (p^1p^2+(p^4)^2)  &  \tfrac12(p^4p^5+p^6p^2) & \tfrac12(p^1p^5+p^4p^6)\\\noalign{\smallskip}
p^4p^6 & p^2p^5 & p^5p^3 & \tfrac12(p^4p^5+p^6p^2) & \tfrac12((p^5)^2+p^2p^3) & \frac12(p^6p^5+p^4p^3)\\\noalign{\smallskip}
p^1p^6 & p^4p^5 & p^6p^3 & \tfrac12(p^1p^5+p^4p^6) & \tfrac12(p^6p^5+p^4p^3) & \tfrac12(p^1p^3+(p^6)^2)
\end{bmatrix} .
\end{equation*}
The determinant of $P$ is $
\rho = \det P = p^1p^2p^3+2p^4p^5p^6-p^1(p^5)^2-p^2(p^6)^2-p^3(p^4)^2,
$. Let $\bd{s}:= [s^1, s^2, s^3, s^4, s^5, s^6]^T = \upsilon(\adj(P))$,
where $\adj(P) = \rho P^{-1}$ is the adjoint matrix of $P$.

The Christoffel symbols are arranged in the following six
symmetric matrices:
\begin{alignat*}{4}
& \Gamma^1 = \frac{-1}{\rho}
\begin{bmatrix}
s^1 & 0 & 0 & s^4 & 0 & s^6 \\
0 & 0 & 0 & 0 & 0 & 0\\
0 & 0 & 0 & 0 & 0 & 0\\
s^4 & 0 & 0 & s^2 & 0 & s^5\\
0 & 0 & 0 & 0 & 0 & 0\\
s^6 & 0 & 0 & s^5 & 0 & s^3 
\end{bmatrix}, & \quad & 
\Gamma^2 = \frac{-1}{\rho}
\begin{bmatrix}
0 & 0 & 0 & 0 & 0 & 0\\
0 & s^2 & 0 & s^4 & s^5 & 0\\
0 & 0 & 0 & 0 & 0 & 0\\
0 & s^4 & 0 & s^1 & s^6 & 0\\
0 & s^5 & 0 & s^6 & s^3 & 0\\
0 & 0 & 0 & 0 & 0 & 0 
\end{bmatrix}, \\\noalign{\medskip}
& \Gamma^3 = \frac{-1}{\rho}
\begin{bmatrix}
0 & 0 & 0 & 0 & 0 & 0\\
0 & 0 & 0 & 0 & 0 & 0\\
0 & 0 & s^3 & 0 & s^5 & s^6\\
0 & 0 & 0 & 0 & 0 & 0\\
0 & 0 & s^5 & 0 & s^2 & s^4\\
0 & 0 & s^6 & 0 & s^4 & s^1 
\end{bmatrix}, & \quad & 
\Gamma^4 = \frac{-1}{2\rho}
\begin{bmatrix}
0 & s^4 & 0 & s^1 & s^6 & 0\\
s^4 & 0 & 0 & s^2 & 0 & s^5\\
0 & 0 & 0 & 0 & 0 & 0\\
s^1 & s^2 & 0 & 2s^4 & s^5 & s^6\\
s^6 & 0 & 0 & s^5 & 0 & s^3\\
0 & s^5 & 0 & s^6 & s^3 & 0 
\end{bmatrix}, \\\noalign{\medskip}
& \Gamma^5 = \frac{-1}{2\rho}
\begin{bmatrix}
0 & 0 & 0 & 0 & 0 & 0\\
0 & 0 & s^5 & 0 & s^2 & s^4\\
0 & s^5 & 0 & s^6 & s^3 & 0\\
0 & 0 & s^6 & 0 & s^4 & s^1\\
0 & s^2 & s^3 & s^4 & 2s^5 & s^6\\
0 & s^4 & 0 & s^1 & s^6 & 0 
\end{bmatrix}, & \quad & 
\Gamma^6 = \frac{-1}{2\rho}
\begin{bmatrix}
0 & 0 & s^6 & 0 & s^4 & s^1\\
0 & 0 & 0 & 0 & 0 & 0\\
s^6 & 0 & 0 & s^5 & 0 & s^3\\
0 & 0 & s^5 & 0 & s^2 & s^4\\
s^4 & 0 & 0 & s^2 & 0 & s^5\\
s^1 & 0 & s^3 & s^4 & s^5 & 2s^6 
\end{bmatrix}.
\end{alignat*}

\end{document}